\documentclass[conference]{IEEEtran}
\usepackage{graphicx}
\usepackage{amsmath, bbm}
\usepackage{amsthm}
\usepackage{amssymb}
\usepackage[font=footnotesize]{subfig}
\usepackage{mathtools}
\usepackage{xcolor}
\usepackage{array,multirow}
\usepackage{setspace}
\usepackage{url}
\usepackage{booktabs}

\DeclareGraphicsExtensions{.pdf,.png,.jpg}
\ifCLASSINFOpdf
\else
\fi
\usepackage{array}
\usepackage{multirow}
\usepackage{booktabs}
\usepackage{adjustbox,lipsum}
\begin{document}\sloppy

\def\x{{\mathbf x}}
\def\L{{\cal L}}

%

\title{Efficient Feature Compression for Machines \\ with Global Statistics Preservation\vspace{-2ex}}

\author{%
Md Eimran Hossain Eimon$^{\dag}$, Hyomin Choi$^{\ast}$, Fabien Racap\'e$^{\ast}$, \\ Mateen Ulhaq$^{\ast}$, Velibor Adzic$^{\dag}$, Hari Kalva$^{\dag}$ and Borko Furht$^{\dag}$\\[0.5em]
{\small\begin{minipage}{\linewidth}\begin{center}
\begin{tabular}{ccc}
$^{\dag}$Florida Atlantic University & \hspace*{0.5in} & $^{\ast}$InterDigital - AI Lab  \\
\url{{meimon2021,vadzic, hkalva, bfurht}@fau.edu} && \url{{hyomin.choi, fabien.racape, mateen.ulhaq}@interdigital.com}
\end{tabular}
\end{center}\end{minipage}}
}


%


\maketitle

\begin{abstract}

The split-inference paradigm divides an artificial intelligence (AI) model into two parts. This necessitates the transfer of intermediate feature data between the two halves. Here, effective compression of the feature data becomes vital.
In this paper, we employ Z-score normalization to efficiently recover the compressed feature data at the decoder side. To examine the efficacy of our method, the proposed method is integrated into the latest Feature Coding for Machines (FCM) codec standard under development by the Moving Picture Experts Group (MPEG). Our method supersedes the existing scaling method used by the current standard under development. It both reduces the overhead bits and improves the end-task accuracy. To further reduce the overhead in certain circumstances, we also propose a simplified method. Experiments show that using our proposed method shows 17.09\% reduction in bitrate on average across different tasks and up to 65.69\% for object tracking without sacrificing the task accuracy.

\end{abstract}


%
\IEEEpeerreviewmaketitle

\section{Introduction}
Artificial intelligence (AI)-enabled smart edge devices are a phenomenon that will bring more AI to many aspects of our lives. The realization of the existing application systems can be categorized into two strategies with respect to the location of the computational workload of deep neural networks (DNNs): on device vs. onto a remote cloud server~\cite{shlezinger2022collaborative}. High-end mobile devices are equipped with AI accelerators that are increasingly becoming more capable at running larger DNNs on captured sensor data. However, not every small-scale device comes with costly accelerators, and are thus limited in hardware resources. In contrast, remote cloud servers have access to more computing and energy resources. Thus, a common strategy is for edge devices to send compressed input to a  more capable server, which then executes a complex DNN model end-to-end. However, attempting to serve too many edge devices may starve the server of resources. Furthermore, the amount of bandwidth required for transmission between the edge devices and the server also incurs a significant cost.

\begin{figure}[t]
    \centering
    \begin{minipage}[b]{1\linewidth}
    \centering
    \includegraphics[width=\textwidth]{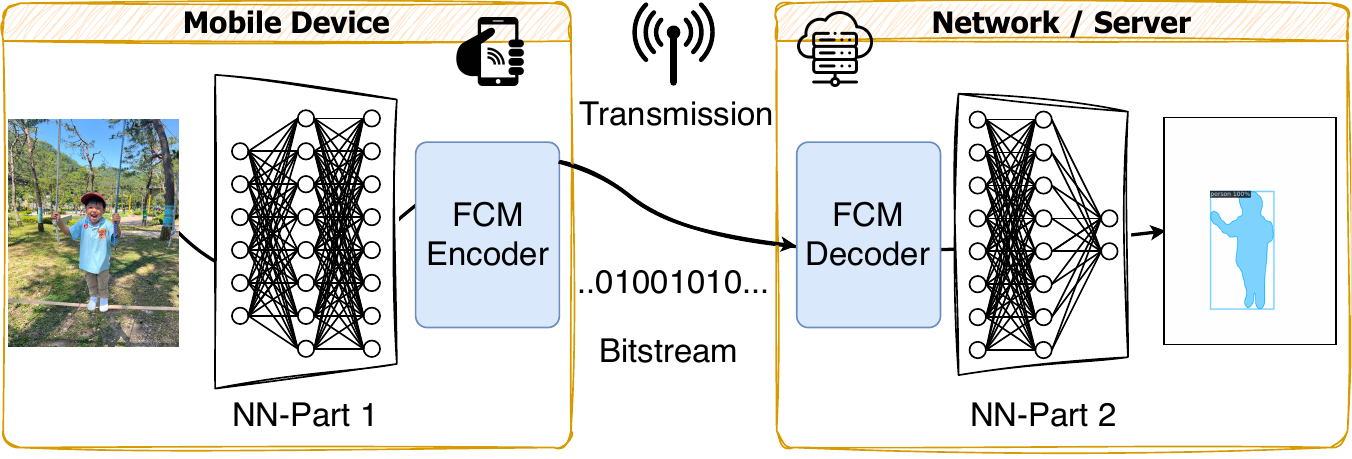}
    \end{minipage}
\caption{An example of the split-inference approach for segmentation with feature coding for machines (FCM) encoder and decoder.}
\vspace{-0.4cm}
\label{fig:split_inference_scenario}
\end{figure}

Split inference (a.k.a. collaborative intelligence) is an emerging approach that achieves a flexible balance between the two strategies. In principle, as depicted in Fig.~\ref{fig:split_inference_scenario}, split inference offers computational-offloading by splitting a DNN into two parts: NN-Part 1 and NN-Part 2 and running them in a collaborative manner. This then necessitates the compression and transmission of feature data instead of coding the captured input. Finding an optimal split point is beyond the scope of the present study, though it is an interesting research topic~\cite{kang2017neurosurgeon}. 

The size of the feature data computed from input varies depending on the split network architecture. It is often much larger than the input size, for instance, at the output of feature pyramid network (FPN) due to the multiple branches of features at different scales~\cite{lin2017feature}. 
This data thus requires significant compression.
Unfortunately, current visual codecs are largely optimized for the human visual system.
Thus, it is crucial that we design codecs optimized for vision analytics.

Recognizing the important role of feature compression for the split inference, the Moving Picture Experts Group (MPEG) established an Ad-hoc group in July 2019 to investigate the needs of the coding for machine vision analytics and initiated the standard development project, called Feature Coding for Machines (FCM). In October 2023, MPEG Working Group 4 (WG4) FCM started with evaluating 12 baseline proposals~\cite{press_release_mpeg144}. According to the latest report~\cite{fcm_cttc}, compared with the remote-inference approach that runs a whole DNN on compressed input, the split inference with FCM achieves the same task accuracy but significantly saves bits up to about 94\%.

The present study introduces a novel scaling method that preserves global statistics of computed features to improve inference task accuracy. Our method periodically signals original statistical parameters to the decoder. Those parameters are used to properly scale the decoded features such that the original global statistics are recovered. By integrating the proposed method to FCM, our method not only improves the task accuracy but also reduces bitrate by superseding the existing scaling method that requires some overhead bits for every frames. Our experiments show that the proposed method saves bits by 17.09\% on average at the same accuracy of multiple tasks on various datasets compared with the latest FCM.

Section~\ref{sec:fcm} briefly reviews overall coding process of FCM in the context of split inference. The proposed method is discussed in Section~\ref{sec:proposed_method}, followed by experimental results in Section~\ref{sec:experiments} and conclusion in Section~\ref{sec:conclusions}.


\section{Feature Coding for Machines}
\label{sec:fcm}

\begin{figure}[t]
    \centering
    \begin{minipage}[b]{1\linewidth}
    \centering
    \includegraphics[width=\textwidth]{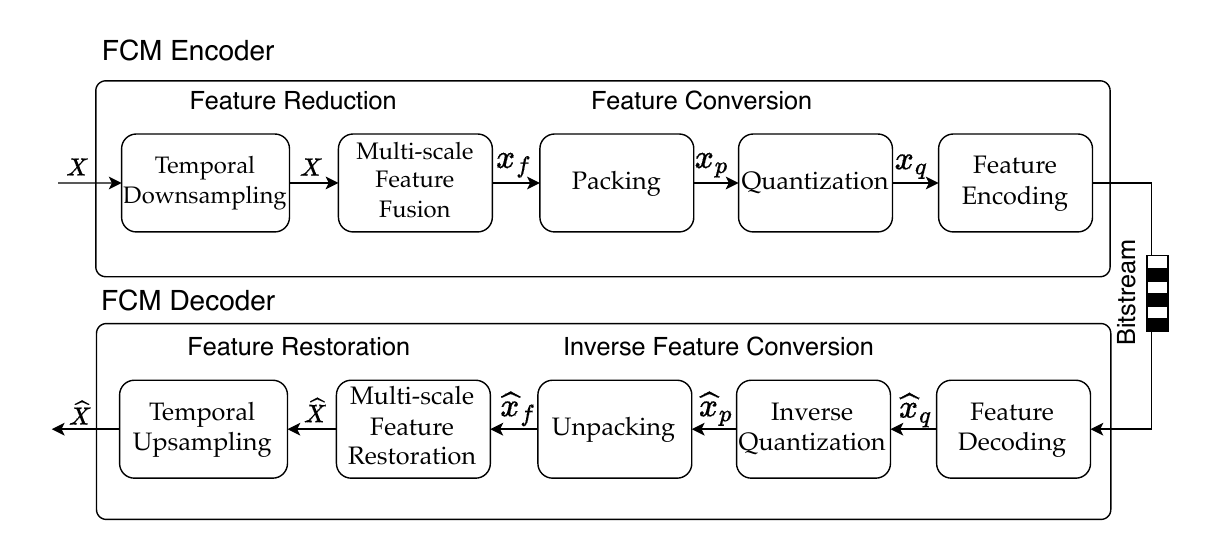}
    \end{minipage}
\caption{A brief overview of FCM codec pipeline.}
\vspace{-0.5cm}
\label{fig:fcm_architecture}
\end{figure}

As shown in Fig.~\ref{fig:split_inference_scenario}, FCM encodes and decodes the intermediate feature data between the partitioned networks: NN-Part 1 and NN-Part 2.
Fig.~\ref{fig:fcm_architecture} presents a brief overview of FCM codec pipeline.
Let $X = \{ \mathbf{x}_n \}_{n=1}^N$ represent a set of $N$ feature tensors computed from an input frame via the NN-Part 1, where $\mathbf{x}_n \in \mathbb{R}^{C_n \times H_n \times W_n}$ is a 3-D feature tensor with size of $C_n$ channels $\times H_n$ height $\times W_n$ width.
First, the FCM encoder can drop every other incoming $X$ if Temporal Downsampling is enabled.
Otherwise, the entire sequence of $X$es continues the feature reduction stage.
Any available $X$ is then fused into a single feature tensor $x_f$ with the NN-based Multi-scale Feature Fusion.
In the feature conversion stage, channels of $x_f$ are tiled and packed into a 2-D frame $x_p$, followed by $q$-bit uniform scalar quantization with minimum and maximum normalization to generate the quantized packed frame $x_q$.
Finally, $x_q$ is encoded into a bitstream with a standard codec, e.g., H.266/VVC~\cite{vvc}.
The FCM decoder takes the bitstream as input and reconstructs the set of feature tensors $\widehat{X}$ by performing the inverse of the above encoding process.

\section{Proposed method}
\label{sec:proposed_method}
\begin{figure}[t]
    \centering
    \begin{minipage}[b]{0.95\linewidth}
    \centering
    \includegraphics[width=\textwidth]{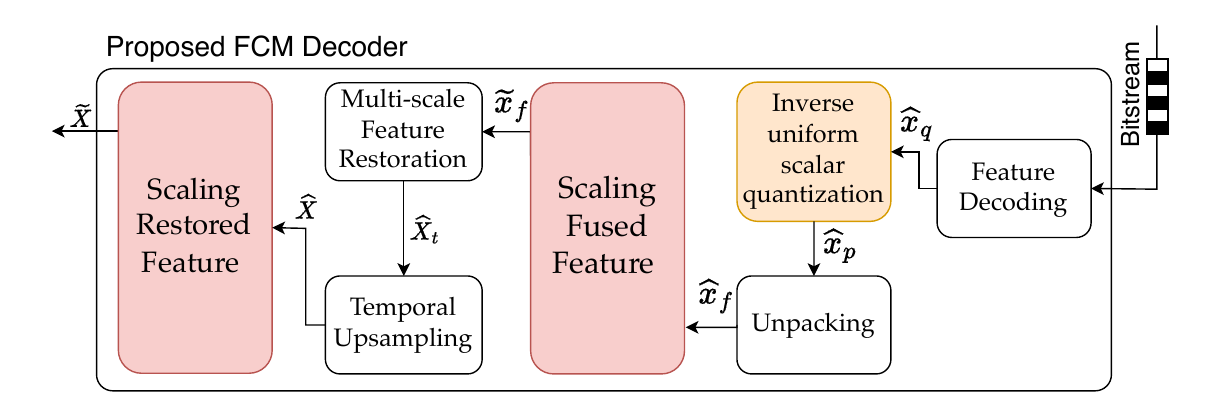}
    \end{minipage}
\caption{Modified FCM decoder with our method for global statistics preservation. Colored blocks are newly added or changed by introducing the proposed method.}
\vspace{-0.4cm}
\label{fig:proposed_architecture}
\end{figure}

Vision inference accuracy appears resilient to hefty amounts of quantization, as long as the decoded intermediate features maintain similar statistical characteristics in the feature space in comparison with the original features~\cite{cohen2021lightweight}.
However, a standard codec, which is employed for the Feature Encoding/Decoding shown in Fig.~\ref{fig:fcm_architecture}, is agnostic to the characteristics of input and downstream tasks. 
As such, due to the inherent nature of lossy compression, the statistical characteristics of reconstructed features often deviate from the original, consequently degrading the task accuracy.
Thus, our method introduces a mean to keep global statistics during the FCM decoding process by scaling the reconstructed features with signaled statistical parameters at two different places as shown in red in Fig.~\ref{fig:proposed_architecture}. 


Inspired by the study~\cite{yang2019wide}, 
if $\mathbf{x}_n$ is taken from the middle of a convolution neural network (CNN),
we may assume that elements $(\mathbf{x}_n)_i \in \mathbf{x}_n$ are distributed as $(\mathbf{x}_n)_i \sim \mathcal{N}(\mu_{\mathbf{x}_n},\,\sigma^{2}_{\mathbf{x}_n})$, where \( \mu_{\mathbf{x}_n} \) and \( \sigma_{\mathbf{x}_n} \) denote the mean and standard deviation of \( \mathbf{x}_n \), respectively.
Similarly, the fused feature tensor $x_f \in \mathbb{R}^{C_f \times H_f \times W_f}$
is also assumed to be distributed as $(\mathbf{x}_f)_i \sim \mathcal{N}(\mu_{\mathbf{x}_f},\,\sigma^{2}_{\mathbf{x}_f})$.

\subsection{Signaling statistical parameters}
Our proposed method requires transmission of the following statistical parameters: $\mu_{\mathbf{x}_n}, \sigma_{\mathbf{x}_n}, \mu_{x_f}, \text{ and } \sigma_{x_f}$.
First, the original statistical parameters of each input feature tensor $\mathbf{x}_n$ are computed as $\mu_{\mathbf{x}_n} = \frac{1}{K_n} \sum_{k=1}^{K_n} \mathbf{x}_n[k]$ and $\sigma_{\mathbf{x}_n} = \sqrt{\frac{1}{K_n} \sum_{k=1}^{K_n} \left( \mathbf{x}_n[k] - \mu_{\mathbf{x}_n} \right)^2}$, where $K_n=C_n H_n W_n$. These are coded into the bitstream. Then, $\mu_{x_f}$ and $\sigma_{x_f}$ of the original fused feature tensor $x_f$ are also computed and transmitted. 

By default, these statistical parameters are coded in the IEEE-754~\cite{ieee_754} single-precision floating-point format (i.e., 32 bits), which costs in total $(N+1) \cdot 2 \cdot 32$ bits per frame.
We also define a refresh period to transmit the parameter updates every $L$ frame, which has a lower overhead than transmitting them every frame.
Our observations indicate that the statistical properties of feature tensors across consecutive frames remain largely similar with marginal variation. 
To leverage this, we synchronize the refresh period $L$ with the intra period in the video encoder. 
Thus, the same statistical parameters computed from the first intra frame are reused for encoding the following $L-1$ frames.

\subsection{Inverse quantization}
At the decoder, a standard codec first parses the input bitstream and reconstructs 2-D packed features $\hat{x}_q$ represented in $q$-bit integer values, where $q \in \{8, 10, 12, ...\}$ depending on specification of the standard codec used in the Feature Encoding/Decoding. Next, inverse uniform scalar quantization is applied to $\hat{x}_q$ to obtain the dequantized features, defined as $\hat{x}_p = \frac{\hat{x}_q}{2^{q} - 1}$. 
Thanks to the proposed method that eventually scales the reconstructed features, the inverse minimum and maximum normalization is skipped in the decoder, reducing a few bytes of signaling minimum and maximum values of features for every frames. Hence, $\hat{x}_p \in [0, 1]$ is directly reshaped into the 3-D feature tensor $\hat{x}_f$ via unpacking.


\subsection{Scaling reconstructed features}
At the decoder, our proposed method uses the received statistical parameters to rescale the reconstructed features $\widehat{X}$ as well as $\hat{x}_f$, as illustrated in Fig.~\ref{fig:proposed_architecture}.
Specifically, we apply Z-score normalization to the incoming tensor, followed immediately by inverse Z-score normalization using the original mean and standard deviation that were computed by the encoder.
This ensures that the reconstructed feature distribution aligns well with the original.

We first rescale the reconstructed fused feature tensor $\hat{x}_f$ with mean $\mu_{\hat{x}_f}=\frac{1}{J} \sum_{j=1}^{J} \hat{x}_f[j]$ and standard deviation $\sigma_{\hat{x}_f} = \sqrt{\frac{1}{J} \sum_{j=1}^{J} \left( \hat{x}_f[j] - \mu_{\hat{x}_f} \right)^2}$, where $J=C_f H_f W_f$.
To do so, we apply Z-score normalization followed by inverse normalization to obtain the rescaled fused feature tensor $\tilde{x}_{f}$, as defined by 
\begin{equation}
\tilde{x}_{f} = \left(\frac{\hat{x}_{f} - {\mu}_{\hat{x}_{f}}}{{\sigma}_{\hat{x}_f}}\right) \times \sigma_{{x}_{f}} +  \mu_{x_f}.
\label{eq:scaling_on_fused_ft}
\end{equation}

Using the rescaled fused feature tensor $\tilde{x}_f$ 
as input, the NN-based Multi-scale Feature Restoration reconstructs the multi-scale feature tensors $\widehat{X} = \{\hat{\mathbf{x}}_{n}\}_{n=1}^N$.
When Temporal Upsampling is enabled, each $\widehat{X}$ is buffered until the next $\widehat{X}$ is reconstructed so that the $\widehat{X}$ between them may be estimated via bilinear interpolation.
The scaling process for $\widehat{X}$ follows a similar process as was done for $\hat{x}_f$ earlier.
That is, the Z-score normalization followed by the inverse of it is applied to each restored feature tensor $\hat{\mathbf{x}}_n$ individually. $\mu_{\hat{\mathbf{x}}_n}$ and $\sigma_{\hat{\mathbf{x}}_n}$ are computed from each $\hat{\mathbf{x}}_n$.
Each rescaled restored feature tensor is then computed by
\begin{equation}
\tilde{\mathbf{x}}_{n} = \left(\frac{\hat{\mathbf{x}}_{n} - {\mu}_{\hat{\mathbf{x}}_{n}}}{{\sigma}_{\hat{\mathbf{x}}_n}}\right) \times \sigma_{\mathbf{x}_{n}} +  \mu_{\mathbf{x}_n}.
\label{eq:scaling_on_restored_ft}
\end{equation}
Finally, the set $\widetilde{X}=\{\tilde{\mathbf{x}}_n\}^{N}_{n=1}$ of rescaled restored feature tensors is fed into NN-Part 2 to accomplish the inference task.

\subsection{Simplified scaling on the restored features}
The previous normalization method transmits global minimum and maximum values every frame, which costs 8 bytes per input frame.
Our proposed normalization method instead transmits $8 (N + 1)$ bytes of statistical parameters for every refresh period of $L$ frames.
Thus, our method necessarily transmits fewer overhead bits if the input video contains more than $N + 1$ frames per refresh period. 
For individual image-based coding, however, the overhead bits required for the proposed method may be burdensome.
Therefore, we further propose a simplified scaling method for the restored features, which especially has efficacy when $N > 1$.

Suppose that $\{\mathbf{x}_n\}^N_{n=1}$ are independent random variables that are distributed as $\mathbf{x}_n \sim \mathcal{N}(\mu_{\mathbf{x}_n}, \sigma^2_{\mathbf{x}_n})$. Since the sum of normally distributed independent random variables is also normally distributed, we can assume that $X \sim \mathcal{N}(\mu_{X}, \sigma^2_{X})$, where 
$\mu_X = \sum_{n=1}^N \mu_{\mathbf{x}_n}$ and $\sigma_X = \sqrt{\sum_{n=1}^N \sigma_{\mathbf{x}_n}^2}$. Therefore, our simplified method codes only a single set of the parameters $(\mu_{X}, \sigma^2_{X})$ for every refresh period of $L$ to compute $\widetilde{X}$ in the decoder. Instead of 32-bit floating-point precision, the simplified method quantizes and codes those parameters in half-precision in the \texttt{bfloat16} format~\cite{bfloat_16}, thereby incurring a cost of only 4 bytes per refresh period.

For the simplified method, we also adjust Eq.~(\ref{eq:scaling_on_restored_ft}) to instead compute each rescaled restored feature tensor $\tilde{\mathbf{x}}_{n}$ as
\begin{equation}
\tilde{\mathbf{x}}_{n} = \left(\frac{\hat{\mathbf{x}}_{n} - {\mu}_{\widehat{X}}}{{\sigma}_{\widehat{X}}}\right) \times \sigma_{X} +  \mu_{X},
\label{eq:scaling_on_restored_ft_with_simplified_method}
\end{equation}
where $\mu_{\widehat{X}} = \sum_{n=1}^{N}\mu_{\hat{\mathbf{x}}_n}$ and $\sigma_{\widehat{X}} = \sqrt{\sum_{n=1}^N \sigma_{\hat{\mathbf{x}}_n}^2}$.




\section{Experimental Results}
\label{sec:experiments}
\begin{table}[t]
\centering
\caption{Summary of the use cases in CTTC~\cite{fcm_cttc}}
\vspace{-0.2cm}
\label{tbl:cttc}
\smallskip\noindent
\resizebox{1.0\linewidth}{!}{%
\fontsize{15pt}{15pt}\selectfont
\begin{tabular}{@{}cc|c|c|c@{}}
\toprule
\multicolumn{2}{c|}{Dataset}           & Task                                                                        & Network                               & Split point (SP)                                                                                 \\ \midrule
\multirow{2}{*}{Image} & OpenImages-v6~\cite{oiv6_seg} & \begin{tabular}[c]{@{}c@{}}Instance\\ Segmentation\end{tabular}             & Mask RCNN-X101-FPN~\cite{mask_rcnn}                    & \multirow{3}{*}{\begin{tabular}[c]{@{}c@{}}4 SPs at \\ Feature\\ Pyramid\\ Network\end{tabular}} \\ \cmidrule(lr){2-4}
                       & OpenImages-v6~\cite{oiv6_det} & \multirow{2}{*}{\begin{tabular}[c]{@{}c@{}}Object\\ Detection\end{tabular}} & \multirow{2}{*}{Faster RCNN-X101-FPN~\cite{faster_rcnn}} &                                                                                                  \\ \cmidrule(r){1-2}
\multirow{3}{*}{Video} & SFU-HW~\cite{sfu_v1} &                                                                             &                                       &                                                                                                  \\ \cmidrule(l){2-5} 
                       & TVD~\cite{tvd}           & \multirow{2}{*}{\begin{tabular}[c]{@{}c@{}}Object\\ Tracking\end{tabular}}  & \multirow{2}{*}{JDE w/ YOLOv3~\cite{wang2019towards}}        & \begin{tabular}[c]{@{}c@{}}3 SPs at\\ Darknet-53~\cite{EEERedmon2018_yolov3}\end{tabular}                                    \\ \cmidrule(lr){2-2} \cmidrule(l){5-5} 
                       & HiEve~\cite{hieve}         &                                                                             &                                       & \begin{tabular}[c]{@{}c@{}}3 SPs near \\ YOLO layers\end{tabular}                                \\ \bottomrule
\end{tabular}}
\vspace{-0.3cm}
\end{table}

Our proposed method is integrated into the FCM Test Model (FCTM)-3.2\footnote{\url{https://git.mpeg.expert/MPEG/Video/fcm/fctm}} and evaluated by following the common test and training conditions (CTTC)~\cite{fcm_cttc} established by MPEG for standards development regarding FCM.
Since the development scope of FCM pertains to coding and decoding the intermediate features at the split point of a vision network, the MPEG FCM group employs CompressAI-Vision\footnote{\url{https://github.com/InterDigitalInc/CompressAI-Vision}} as a common evaluation platform.
This platform manages the use cases of various vision networks with several split points and evaluates the performance of inference from the reconstructed features by the FCTM. 

As summarized in Table~\ref{tbl:cttc}, the CTTC~\cite{fcm_cttc} designates five datasets for images and videos, which are used by three different vision networks along with predefined split points.
For the various image-based tasks, 5000 images are selected from OpenImages-v6 (OIV6) according to~\cite{fcm_cttc}. For the object detection and tracking tasks on video, there are three video datasets, one of which includes several video sequences of various lengths.
Following these uses cases, the FCTM-3.2 with the CTTC has been evaluated and used as benchmark.
For feature coding within the FCTM-3.2, the VVC Test model (VTM)-12.0~\footnote{\url{https://vcgit.hhi.fraunhofer.de/jvet/VVCSoftware_VTM}} is employed with All Intra and low-delay B configuration for the image and video datasets, respectively. 

\subsection{Split points}
Depending on the vision networks associated with the end task and datasets as shown in Table~\ref{tbl:cttc}, there are different split points defined.
For the Faster RCNN-X101-FPN~\cite{faster_rcnn} and Mask RCNN-X101-FPN~\cite{mask_rcnn} models, the CTTC~\cite{fcm_cttc} splits at the feature pyramid network (FPN), thereby generating four feature tensors $X = \{ \mathbf{x}_n \}_{n=1}^4$ with the same number of channels $C_n = 256$ and a channel resolution of $H_n \times W_n = \frac{H_r}{2^{(n+1)}} \times \frac{W_r}{2^{(n+1)}}$, where $H_r \times W_r$ is the scaled input resolution to the vision networks.
For the tracking task, we use the YOLOv3-based JDE~\cite{wang2019towards} and employ two different split points that generate three feature tensors $X = \{ \mathbf{x}_n \}_{n=1}^3$.
For the first split use case at the backbone of Darknet~\cite{EEERedmon2018_yolov3}, we have feature tensors with $\mathbf{x}_1 \in \mathbb{R}^{256 \times 76 \times 136}$, $\mathbf{x}_2 \in \mathbb{R}^{512 \times 38 \times 68}$, and $\mathbf{x}_3 \in \mathbb{R}^{1024 \times 19 \times 34}$.
The other split use case generates the feature tensors with various resolutions of ${128 \times 76 \times 136}$, ${256 \times 38 \times 68}$, and ${512 \times 19 \times 34}$.
More details on the split use cases associated with the task networks can be found in~\cite{fcm_cttc}.

\subsection{Evaluation}

\begin{table}[t]
\centering
\caption{Summarized BD-rate-accuracy reduction by our methods against FCTM-3.2 with CTTC~\cite{fcm_cttc}}
\label{tbl:bd-rate-fctm}
\resizebox{0.95\linewidth}{!}{%
\fontsize{13pt}{13pt}\selectfont
\begin{tabular}{@{}cccc@{}}
\toprule
Task                                                                        & Dataset         & Ours  & Ours (Simplified)    \\ \midrule
\begin{tabular}[c]{@{}c@{}}Instance\\ Segmentation\end{tabular}             & OpenImages-v6   & 13.15\% & 4.86\%  \\ \midrule
\multirow{4}{*}{\begin{tabular}[c]{@{}c@{}}Object\\ Detection\end{tabular}} & OpenImages-v6   & 4.20\% & \textbf{-10.53}\% \\
                                                                            & SFU-HW-ClassA/B & \textbf{-6.14}\% & -4.84\%  \\
                                                                            & SFU-HW-ClassC   & 4.63\% & 4.89\%   \\
                                                                            & SFU-HW-Class D  & -3.23\% & \textbf{-3.84}\%  \\ \midrule
\multirow{3}{*}{\begin{tabular}[c]{@{}c@{}}Object\\ Tracking\end{tabular}}  & TVD             & \textbf{-18.76}\% & -13.95\% \\
                                                                            & HiEve-1080p   & -59.64\% & \textbf{-65.69}\% \\
                                                                            & HiEve-720p    & -39.03\% & \textbf{-47.65}\% \\ \midrule
\multicolumn{2}{c}{Overall}                                                                   & -13.10\% & \textbf{-17.09}\% \\ \bottomrule
\end{tabular}}
\end{table}

To evaluate coding efficiency, we employ BD-Rate~\cite{bd_rate} and measure bitrate vs task accuracy. Which accuracy metric is used depends on the task and dataset. For example, mean average precision (mAP) is used for object detection and instance segmentation, whereas multiple object tracking accuracy (MOTA) is employed for object tracking. 

Table~\ref{tbl:bd-rate-fctm} summarizes overall coding gain of FCTM-3.2 adopting our methods against the default FCTM-3.2.
The results show that our method achieves significant reductions in bitrate on certain datasets.
In particular, the tracking task shows an especially large reduction of 65.69\% on the HiEve-1080p dataset with our simplified method.
We attribute this improvement to our method's ability to maintain statistical consistency between consecutive frames, which is especially important for tracking.
In contrast, spatial statistics are more important than temporal statistics for the object detection and segmentation task.
Although the coding gain for those tasks is smaller than for the tracking task, our method still shows bitrate reduction of up to 6.14\% for the object detection on the video datasets.
For the image-based segmentation task, the coding gain with our simplified method is marginally inferior to the benchmark.
Nonetheless, our simplified method still holds promise for still images and works better in this domain than our primary method, which is more video-oriented.

\begin{figure}[t]
\centering
    \begin{minipage}[b]{0.49\linewidth}
    \centering
    \includegraphics[width=\textwidth]{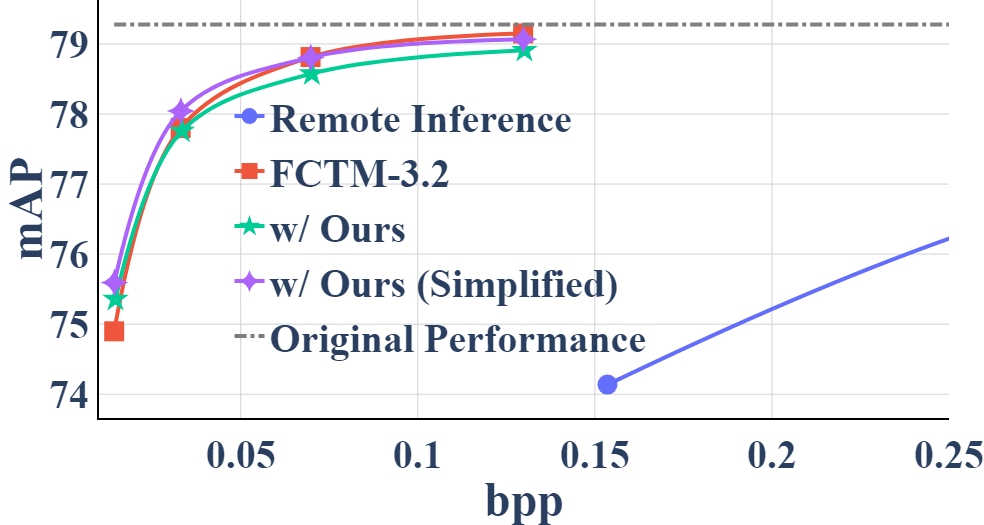}
    \centerline{\footnotesize (a) Object Detection on OIV6}
    \end{minipage}
    \hfill
    \begin{minipage}[b]{0.49\linewidth}
    \centering
    \includegraphics[width=\textwidth]{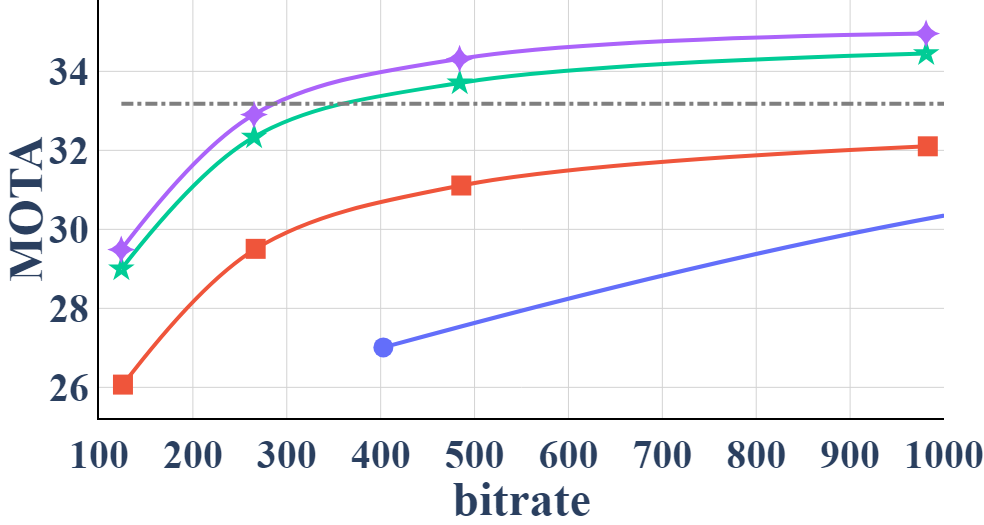}
    \centerline{\footnotesize (b) Object Tracking on HiEve-1080p}
    \end{minipage}
\caption{Comparison of rate-accuracy curves between various coding methods on different tasks. Note, the legend is shared.}
\label{fig:rd_plots}
\end{figure}

Fig.~\ref{fig:rd_plots} compares rate-accuracy curves of the FCTM-3.2 with our methods, denoted by ``w/ Ours (Simplified),'' against the default FCTM-3.2 in red and ``Remote Inference'' in blue. The remote inference is the scenario that compresses and decompress input image/video, for example using VTM-12.0, from which the whole vision network performs inference on the reconstructed pixels. Additionally, the dashed bar shows the original inference accuracy when the task networks are examined without split or compression. In both Fig.~\ref{fig:rd_plots}(a) and (b) on different tasks and datasets, overall feature compression demonstrates an undeniable coding gain against remote inference. Moreover, our methods further improve the coding efficiency on top of the FCTM-3.2 in terms of rate-accuracy.

\begin{figure}[t]
\centering
    \begin{minipage}[b]{0.32\linewidth}
    \centering
    \centerline{\small \textcolor{white}{g}Ground Truth\textcolor{white}{g}}
    \vspace{0.1cm}
    \includegraphics[width=\textwidth]{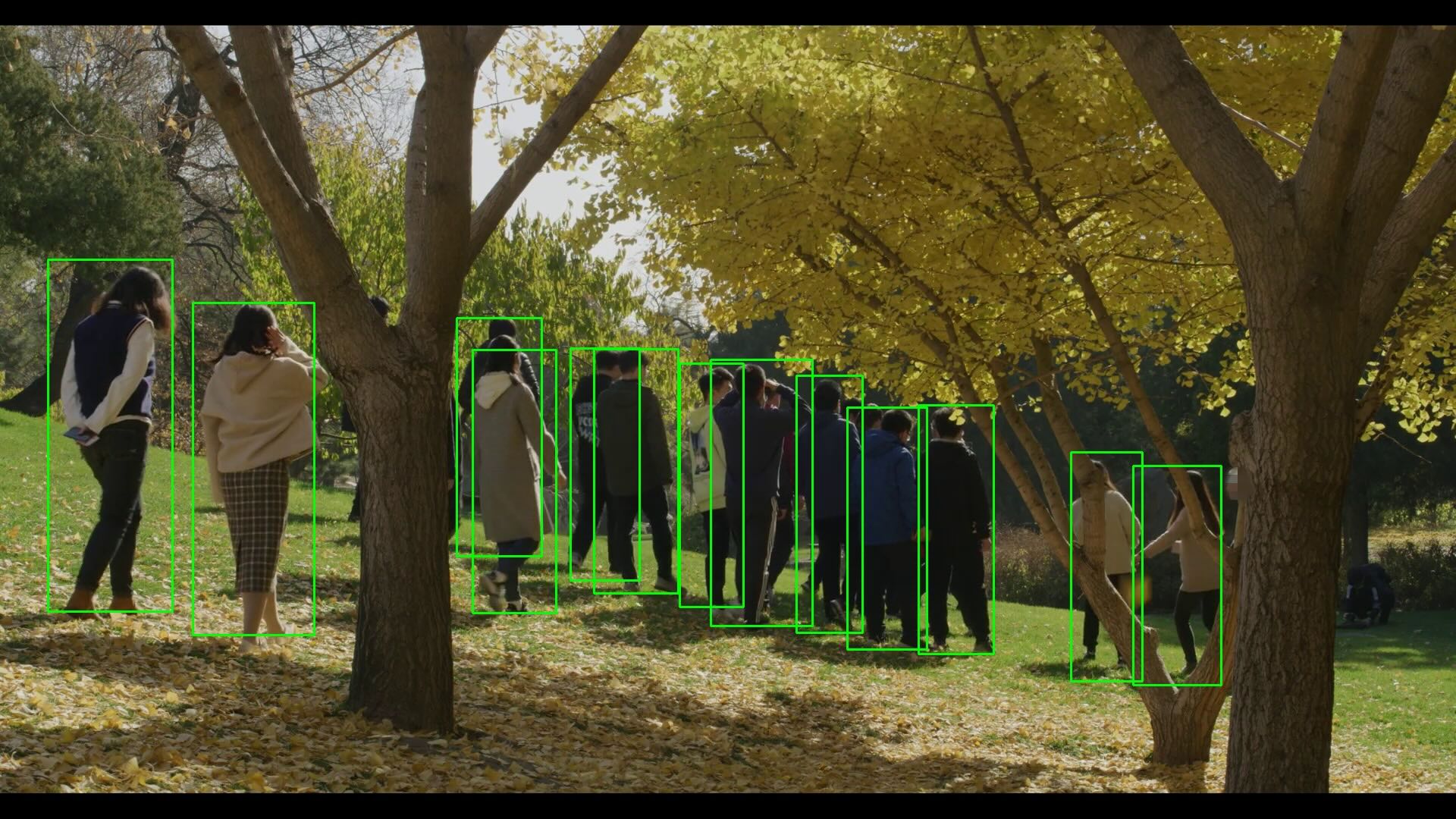}
    \centerline{\footnotesize TVD-03 (\#720th)}\medskip
    \end{minipage}
    \begin{minipage}[b]{0.32\linewidth}
    \centering
    \centerline{\small \textcolor{white}{g}FCTM-3.2\textcolor{white}{g}}
    \vspace{0.1cm}
    \includegraphics[width=\textwidth]{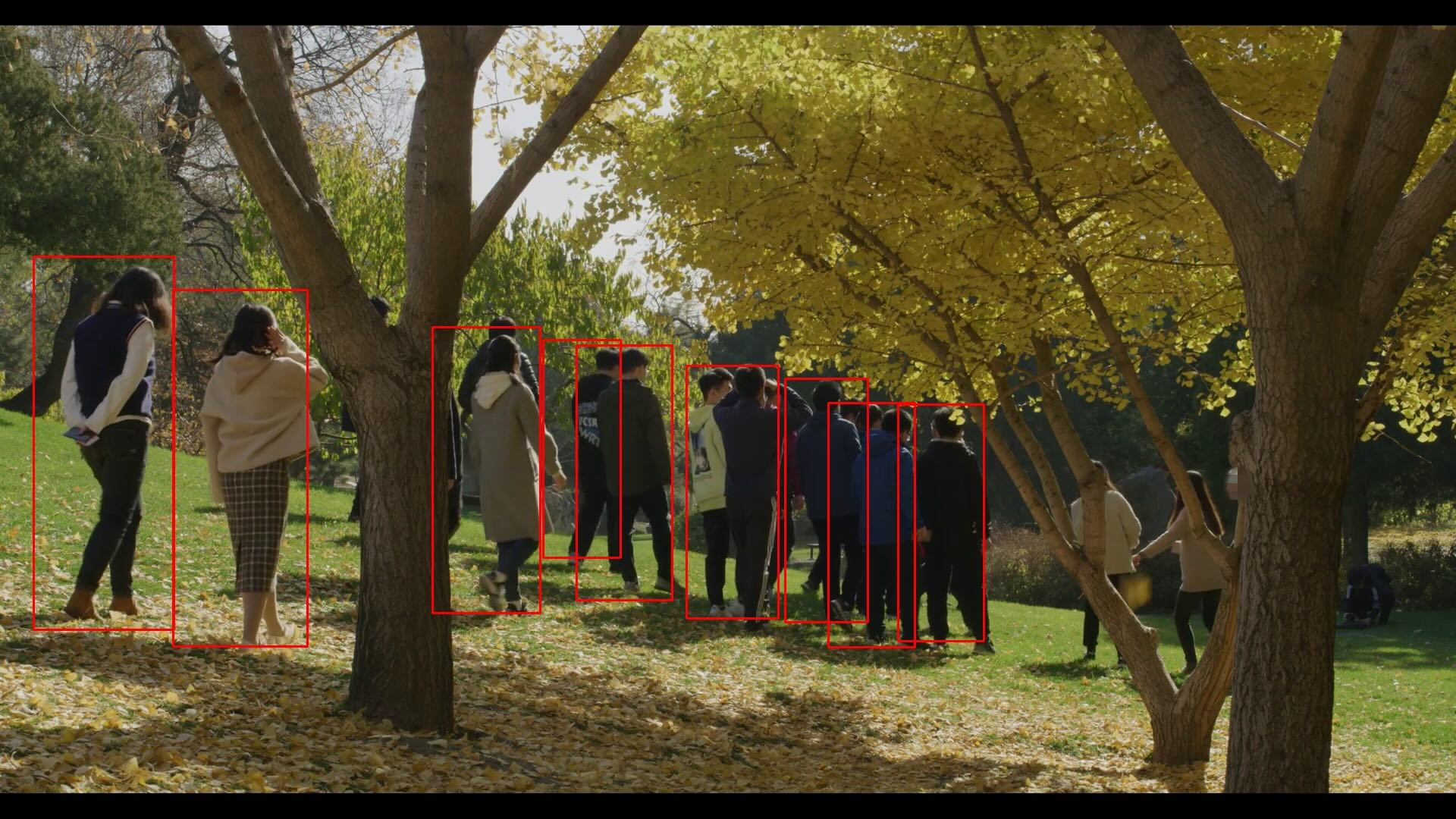}
    \centerline{\footnotesize [63.915, 60.62]}\medskip
    \end{minipage}
    \begin{minipage}[b]{0.32\linewidth}
    \centering
    \centerline{\small \textcolor{white}{g}Ours (Simplified)\textcolor{white}{g}}
    \vspace{0.1cm}
    \includegraphics[width=\textwidth]{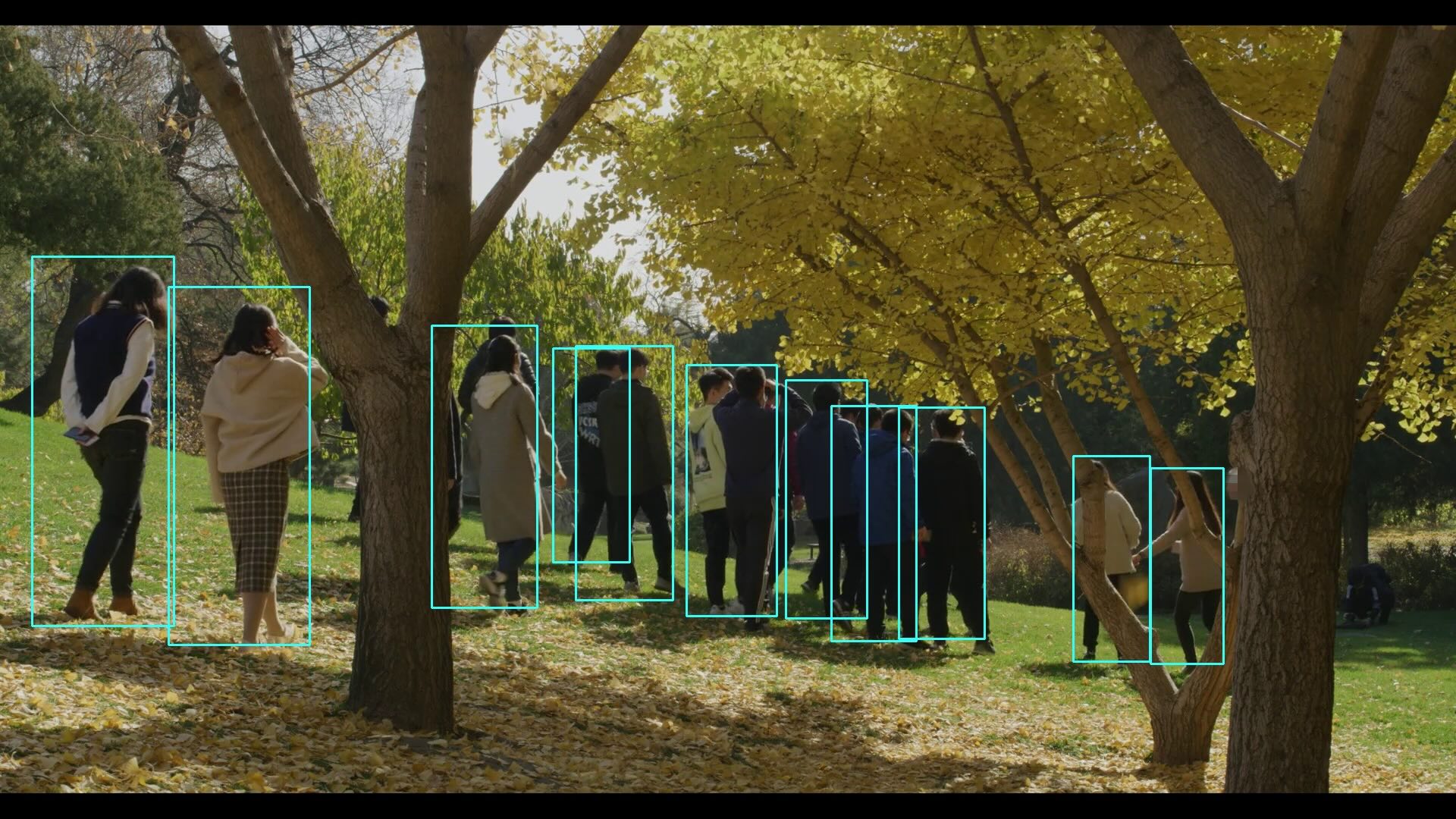}
    \centerline{\footnotesize [\textbf{60.792, 64.44}]}\medskip
    \end{minipage}
    
    \centering
    \begin{minipage}[b]{0.32\linewidth}
    \centering
    \includegraphics[width=\textwidth]{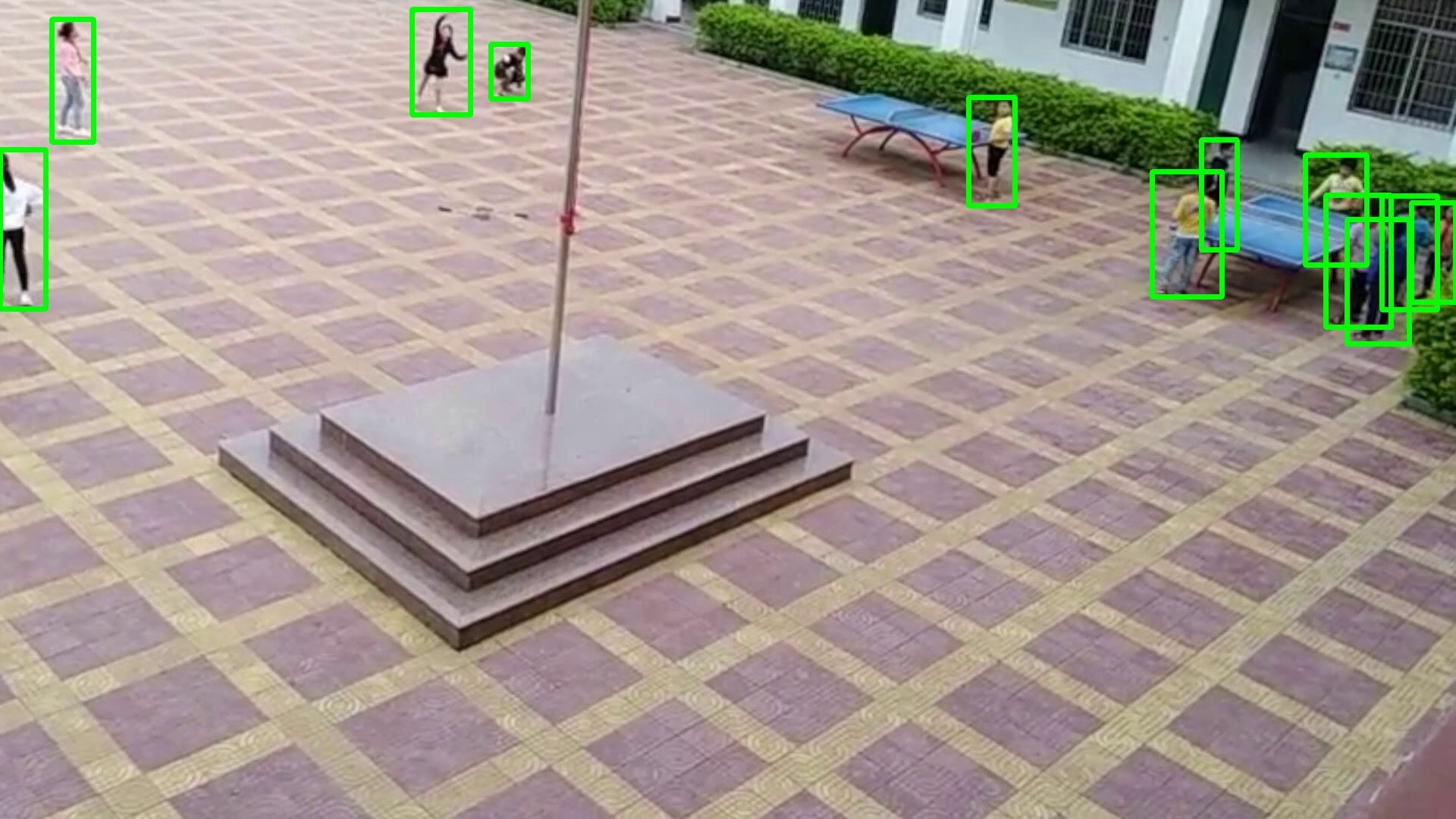}
    \centerline{\footnotesize HiEve-13 (\#97th)}\medskip
    \end{minipage}
    \begin{minipage}[b]{0.32\linewidth}
    \centering
    \includegraphics[width=\textwidth]{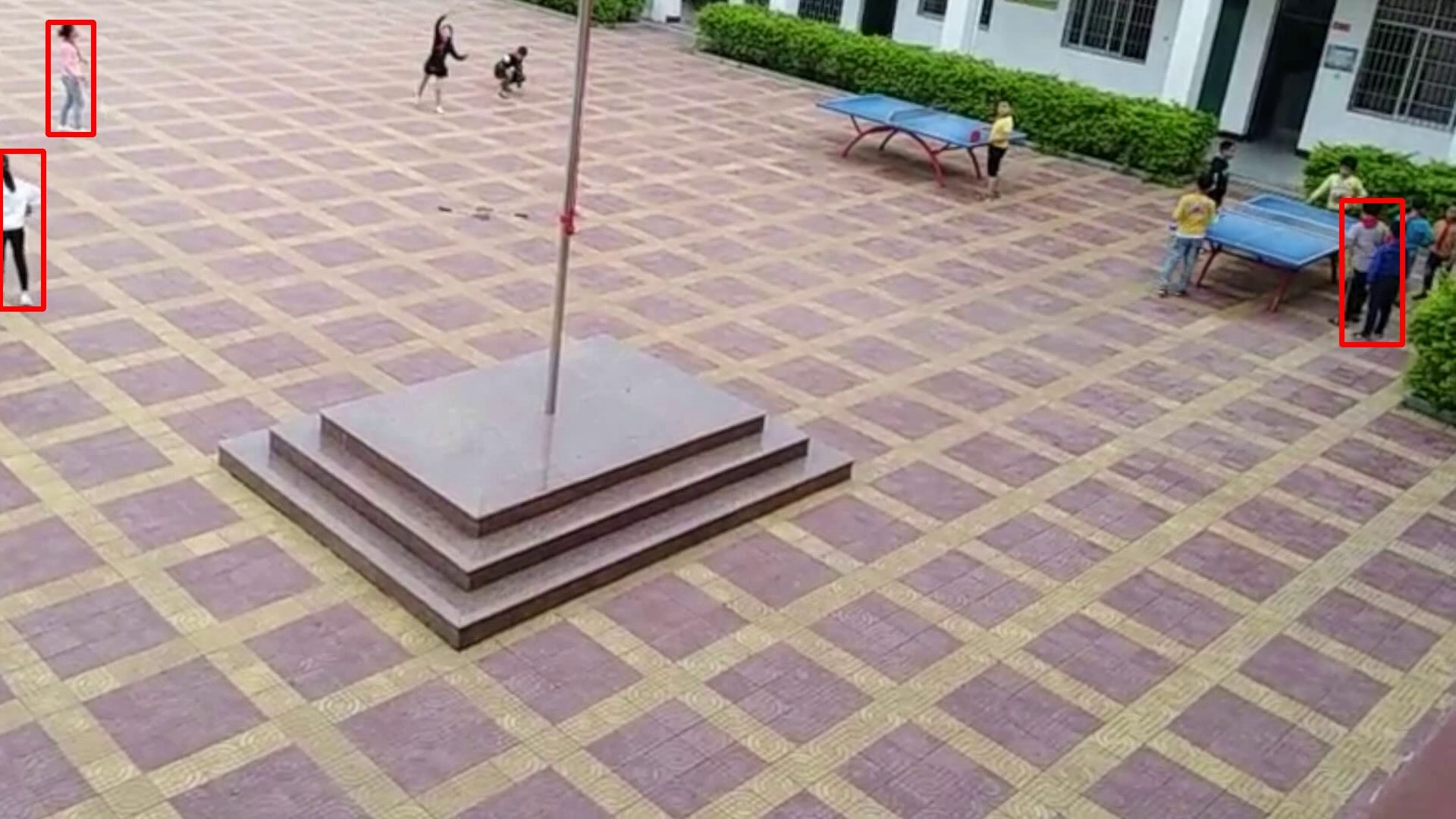}
    \centerline{\footnotesize [503.143, 53.45]}\medskip
    \end{minipage}
    \begin{minipage}[b]{0.32\linewidth}
    \centering
    \includegraphics[width=\textwidth]{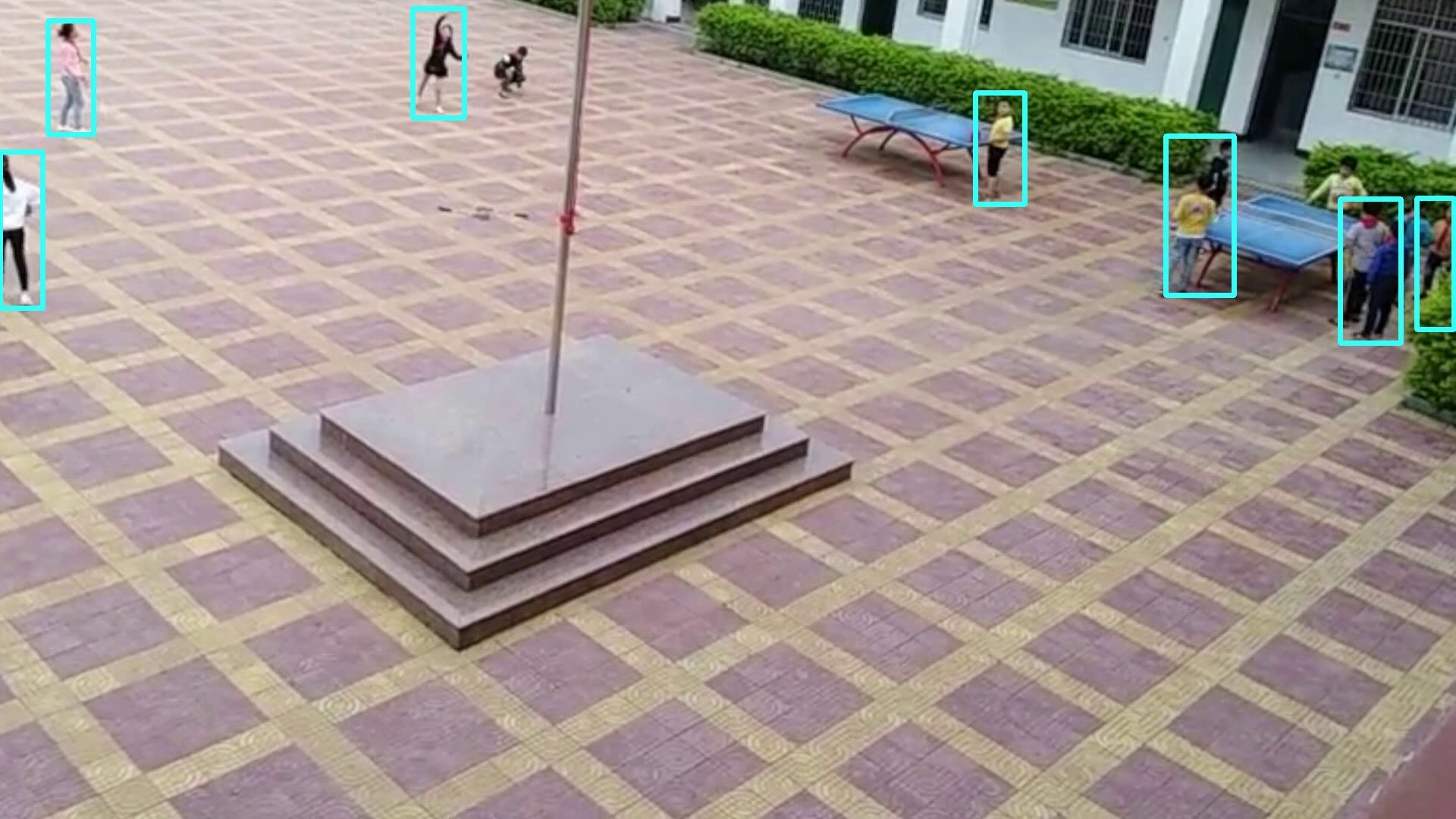}
    \centerline{\footnotesize [\textbf{501.315, 59.51}]}\medskip
    \end{minipage}
    
    \centering
    \begin{minipage}[b]{0.32\linewidth}
    \centering
    \includegraphics[width=\textwidth]{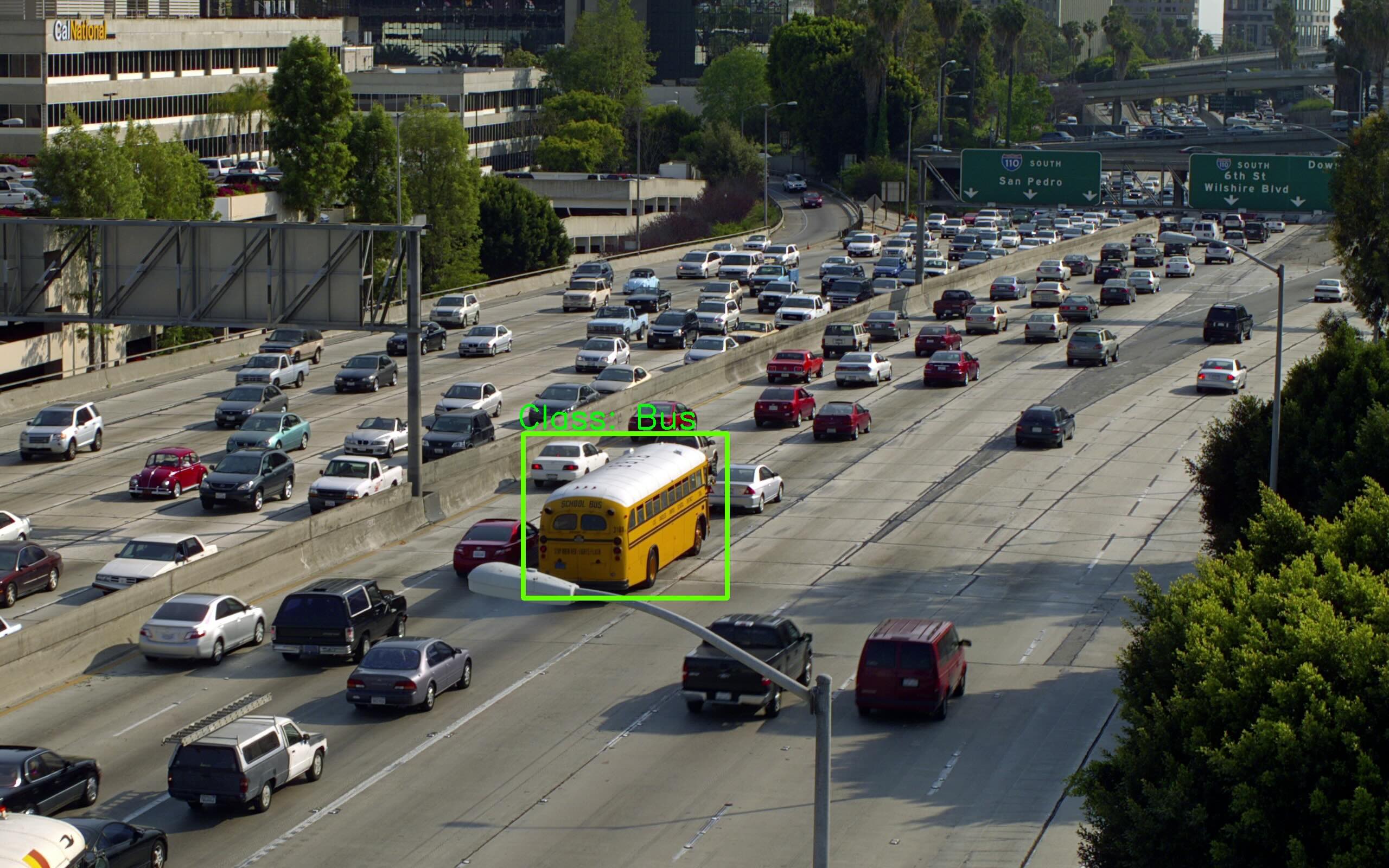}
    \centerline{\footnotesize Traffic (\#31st)}\medskip
    \end{minipage}
    \begin{minipage}[b]{0.32\linewidth}
    \centering
    \includegraphics[width=\textwidth]{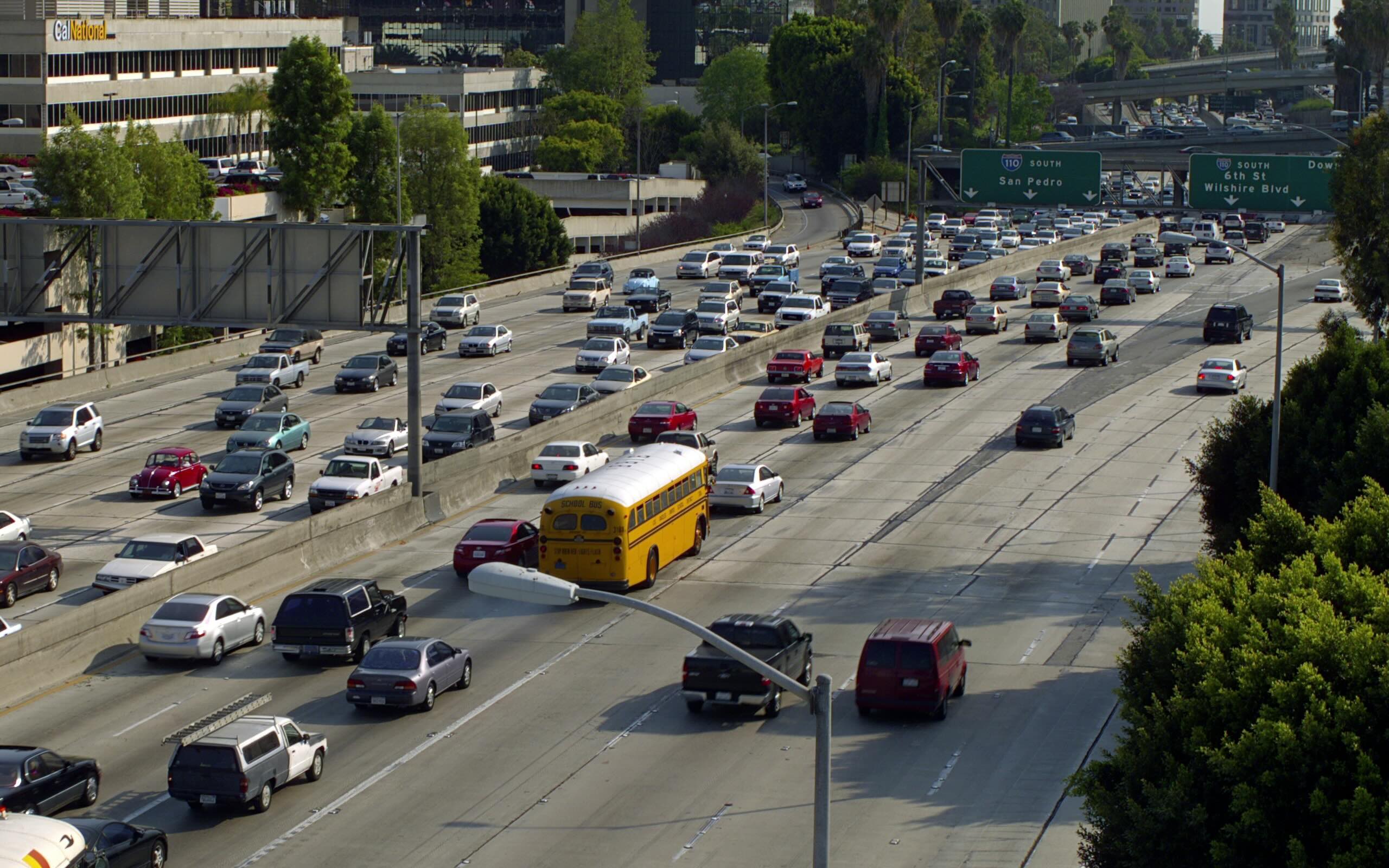}
    \centerline{\footnotesize [38.422, 13.94]}\medskip
    \end{minipage}
    \begin{minipage}[b]{0.32\linewidth}
    \centering
    \includegraphics[width=\textwidth]{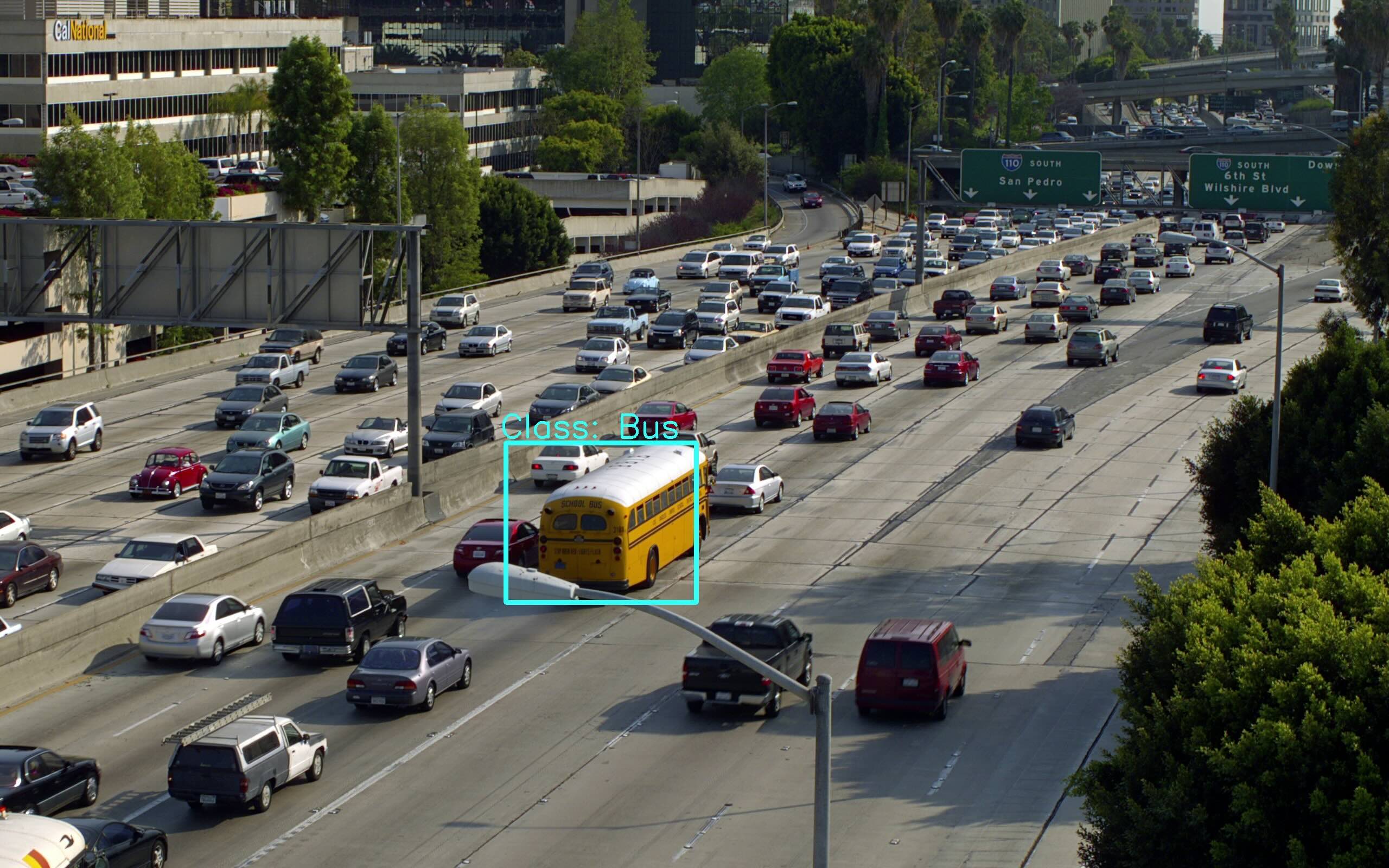}
    \centerline{\footnotesize [\textbf{36.691, 35.31}]}\medskip
    \end{minipage}
\caption{Visual comparison of inference results overlaid on the input frame. First column shows the ground truth. The second and third demonstrate the inference results with compressed feature with FCTM-3.2 and ours, respectively. First two rows present tracking outputs on \textit{TVD-03} and \textit{HiEve-13} at the sample frame with the number in the bracket, respectively. The last row shows the detected bus on a video sequence, \textit{Traffic}. Overall coding performance for each sequence is presented in [kbps, MOTA/mAP].}
\label{fig:visual_outcomes}
\end{figure}

\subsection{Visual comparison}
Fig. \ref{fig:visual_outcomes} visually compares the inference outcomes between the ground truth in green, with FCTM-3.2 in red, and with our proposed method in cyan, all of which are overlaid on the corresponding input frames. Evidently, the first row shows that the model using FCTM-3.2 for feature coding failed to detect the two rightmost individuals, leading to a reduced MOTA. In contrast, our method achieves higher MOTA by 64.44\% by allowing the detection of most of the people while saving bits by 4.89\%. Similarly, in the second row, FCTM-3.2 allows detection of only 3 people in the scene. Meanwhile, with our methods, 7 people out of 13 are detected, resulting in about 6\% increase in MOTA with a few bit savings. Lastly in the third row for the object detection task\textemdash only the bus is highlighted for visual simplicity\textemdash FCTM-3.2 fails to allow detection of any buses. However, our proposed method allowed the bus to be detected, increasing the mAP from 13.94 to 35.31 while saving bits by 4.50\% across the sequence.

\section{Conclusions}
\label{sec:conclusions}
We introduced a Z-score normalization-based scaling method to the latest FCM codec that compresses and transmits the intermediate features in the context of split inference. Our method preserves global statistics of reconstructed features at the decoder by transmitting the original statistical parameters, helping maintain end-task accuracy despite compression noise. We also presented a simplified signaling mechanism to further reduce overhead bits for transmitting the parameters. By integrating our method, FCTM-3.2 saves 17.09\% more bits on average at the same task accuracy across different datasets.





%
\newpage
\bibliographystyle{IEEEbib_abbrev}
\bibliography{ref}

@article{bd_rate,
  title={Calculation of average PSNR differences between RD-curves},
  author={Bjontegaard, G},
  journal={ITU-T SG16 Q},
  volume={6},
  year={2001}
}

@misc{bfloat_16,
	author={Intel},
	title={{BFLOAT16}–Hardware Numerics Definition White Paper},
	howpublished={\url{https://www.intel.com/content/dam/develop/external/us/en/documents/bf16-hardware-numerics-definition-white-paper.pdf}}, 
	year={2018},
	month={Nov.}
}

@article{ieee_754,
	author={{Microprocessor Standards Committee of the IEEE Computer Society}},
	title={{IEEE} standard for foating-point arithmetic},
	journal={IEEE Std 754-2008}, 
	year={2008},
	month={Aug.}
}

@inproceedings {fcm_cttc,
	author={{ISO/IEC JTC 1/SC 29/WG 04}},
	title={Common test and training conditions for {FCM}},
	booktitle={ISO/IEC JTC 1/SC 29/WG 04 {[N0548]}},
    year={2024},
	month={July}
}

@article{sfu_v1,
    title = {A dataset of labelled objects on raw video sequences},
    journal = {Data in Brief},
    volume = {34},
    pages = {106701},
    year = {2021},
    issn = {2352-3409},
    doi = {https://doi.org/10.1016/j.dib.2020.106701},
    url = {https://www.sciencedirect.com/science/article/pii/S2352340920315808},
    author = {Hyomin Choi and Elahe Hosseini and Saeed {Ranjbar Alvar} and Robert A. Cohen and Ivan V. Bajić}

}

@INPROCEEDINGS{tvd,
  author={Gao, Wen and Xu, Xiaozhong and Qin, Matthew and Liu, Shan},
  booktitle={2022 IEEE International Conference on Image Processing (ICIP)}, 
  title={An Open Dataset for Video Coding for Machines Standardization}, 
  year={2022},
  volume={},
  number={},
  pages={4008-4012},
  doi={10.1109/ICIP46576.2022.9897525}}

@article{hieve,
  author       = {Weiyao Lin and
                  Huabin Liu and
                  Shizhan Liu and
                  Yuxi Li and
                  Guo{-}Jun Qi and
                  Rui Qian and
                  Tao Wang and
                  Nicu Sebe and
                  Ning Xu and
                  Hongkai Xiong and
                  Mubarak Shah},
  title        = {Human in Events: {A} Large-Scale Benchmark for Human-centric Video
                  Analysis in Complex Events},
  journal      = {CoRR},
  volume       = {abs/2005.04490},
  year         = {2020},
  url          = {https://arxiv.org/abs/2005.04490},
  eprinttype    = {arXiv},
  eprint       = {2005.04490},
  bibsource    = {dblp computer science bibliography, https://dblp.org}
}

@article{oiv6_det,
  author = {Alina Kuznetsova and Hassan Rom and Neil Alldrin and Jasper Uijlings and Ivan Krasin and Jordi Pont-Tuset and Shahab Kamali and Stefan Popov and Matteo Malloci and Alexander Kolesnikov and Tom Duerig and Vittorio Ferrari},
  title = {The Open Images Dataset V4: Unified image classification, object detection, and visual relationship detection at scale},
  year = {2020},
  journal = {IJCV}
}

@inproceedings{oiv6_seg,
  author = {Rodrigo Benenson and Stefan Popov and Vittorio Ferrari},
  title = {Large-scale interactive object segmentation with human annotators},
  booktitle = {CVPR},
  year = {2019}
}

@article{mask_rcnn,
  author       = {Kaiming He and
                  Georgia Gkioxari and
                  Piotr Doll{\'{a}}r and
                  Ross B. Girshick},
  title        = {Mask {R-CNN}},
  journal      = {CoRR},
  volume       = {abs/1703.06870},
  year         = {2017},
  url          = {http://arxiv.org/abs/1703.06870},
  eprinttype    = {arXiv},
  eprint       = {1703.06870},
  bibsource    = {dblp computer science bibliography, https://dblp.org}
}

@article{faster_rcnn,
  author       = {Shaoqing Ren and
                  Kaiming He and
                  Ross B. Girshick and
                  Jian Sun},
  title        = {Faster {R-CNN:} Towards Real-Time Object Detection with Region Proposal
                  Networks},
  journal      = {CoRR},
  volume       = {abs/1506.01497},
  year         = {2015},
  url          = {http://arxiv.org/abs/1506.01497},
  eprinttype    = {arXiv},
  eprint       = {1506.01497},
  bibsource    = {dblp computer science bibliography, https://dblp.org}
}

@ARTICLE{vvc,
  author={Bross, Benjamin and Wang, Ye-Kui and Ye, Yan and Liu, Shan and Chen, Jianle and Sullivan, Gary J. and Ohm, Jens-Rainer},
  journal={IEEE Transactions on Circuits and Systems for Video Technology}, 
  title={Overview of the Versatile Video Coding {(VVC)} Standard and its Applications}, 
  year={2021},
  volume={31},
  number={10},
  pages={3736-3764},
  doi={10.1109/TCSVT.2021.3101953}}

@article{shlezinger2022collaborative,
  title={Collaborative inference for {AI}-empowered {IoT} devices},
  author={Shlezinger, Nir and Baji{\'c}, Ivan V},
  journal={IEEE Internet of Things Magazine},
  volume={5},
  number={4},
  pages={92--98},
  year={2022},
  publisher={IEEE}
}

@article{kang2017neurosurgeon,
  title={Neurosurgeon: Collaborative intelligence between the cloud and mobile edge},
  author={Kang, Yiping and Hauswald, Johann and Gao, Cao and Rovinski, Austin and Mudge, Trevor and Mars, Jason and Tang, Lingjia},
  journal={ACM SIGARCH Computer Architecture News},
  volume={45},
  number={1},
  pages={615--629},
  year={2017},
  publisher={ACM New York, NY, USA}
}

@inproceedings{lin2017feature,
  title={Feature pyramid networks for object detection},
  author={Lin, Tsung-Yi and Doll{\'a}r, Piotr and Girshick, Ross and He, Kaiming and Hariharan, Bharath and Belongie, Serge},
  booktitle={Proceedings of the IEEE conference on computer vision and pattern recognition},
  pages={2117--2125},
  year={2017}
}

@inproceedings {press_release_mpeg144,
	author={{ISO/IEC JTC 1/SC 29/AG 03}},
	title={Press release of 144th {MPEG} meeting},
	booktitle={ISO/IEC JTC 1/SC 29/AG 03 {[N0129]}},
    year={2023},
	month={October}
}

@article{yang2019wide,
  title={Wide feedforward or recurrent neural networks of any architecture are gaussian processes},
  author={Yang, Greg},
  journal={Advances in Neural Information Processing Systems},
  volume={32},
  year={2019}
}

@article{cohen2021lightweight,
  title={Lightweight compression of intermediate neural network features for collaborative intelligence},
  author={Cohen, Robert A and Choi, Hyomin and Baji{\'c}, Ivan V},
  journal={IEEE Open Journal of Circuits and Systems},
  volume={2},
  pages={350--362},
  year={2021},
  publisher={IEEE}
}

@article{wang2019towards,
  title={Towards Real-Time Multi-Object Tracking},
  author={Wang, Zhongdao and Zheng, Liang and Liu, Yixuan and Wang, Shengjin},
  journal={The European Conference on Computer Vision (ECCV)},
  year={2020}
}

@article{EEERedmon2018_yolov3,
  author    = {Joseph Redmon and
               Ali Farhadi},
  title     = {{YOLOv3}: An Incremental Improvement},
  journal   = {CoRR},
  volume    = {abs/1804.02767},
  year      = {2018},
  url       = {http://arxiv.org/abs/1804.02767},
  archivePrefix = {arXiv},
  eprint    = {1804.02767},
  bibsource = {dblp computer science bibliography, https://dblp.org}
}

\end{document}